\documentclass[10pt,twocolumn,letterpaper]{article}
\usepackage{cvpr}

\usepackage[dvipsnames]{xcolor}

\usepackage{floatrow}
\usepackage{esvect}
\usepackage{xcolor}
\usepackage{booktabs}
\usepackage{multicol}
\usepackage{float}
\usepackage{graphicx}
\usepackage{tabularx}
\usepackage{pgfplots}
\usepackage{comment}
\usepackage{stfloats}
\usepackage[tableposition=top]{caption}
\newcommand\myeq{\mkern1.5mu{=}\mkern1.5mu} 
\usepackage{tikz}
\usepackage{xparse}
\pgfplotsset{compat=1.13,
    /pgfplots/ybar legend/.style={
    /pgfplots/legend image code/.code={%
       \draw[##1,/tikz/.cd,yshift=-0.25em]
        (0cm,0cm) rectangle (3pt,0.8em);},
   },
}
\usepackage{svg}
\usepackage{amsmath}
\usepackage{amsthm}
\usepackage{amsfonts}
\usepackage{amssymb}
\usepackage{bm}
\usepackage{booktabs}
\usepackage{multirow}
\usepackage{enumitem}
\usepackage{mdframed,lipsum}
\usepackage{svg}

\DeclareMathOperator*{\argmin}{arg\,min}
\usepackage[accsupp]{axessibility}
\makeatletter
\apptocmd\@maketitle{{\myfigure{}\par}}{}{}
\makeatother
\usepackage{capt-of,etoolbox}

\definecolor{cvprblue}{rgb}{0.21,0.49,0.74}

\usepackage[pagebackref,breaklinks,colorlinks,citecolor=cvprblue]{hyperref}

\usepackage[capitalize]{cleveref}
\crefname{section}{Sec.}{Secs.}
\Crefname{section}{Section}{Sections}
\Crefname{table}{Table}{Tables}
\crefname{table}{Tab.}{Tabs.}

\def\confName{CVPR}
\def\confYear{2024}

\title{Doodle Your 3D: From \textit{Abstract} Freehand Sketches to \textit{Precise} 3D Shapes}

\author{\\[-1.05cm] \href{https://hmrishavbandy.github.io/}{Hmrishav Bandyopadhyay}\textsuperscript{1} \hspace{.2cm} \href{https://subhadeepkoley.github.io/}{Subhadeep Koley}\textsuperscript{1,2} \hspace{.2cm} \href{https://ayandas.me/}{Ayan Das}\textsuperscript{1} \hspace{.2cm}  \href{https://ayankumarbhunia.github.io/}{Ayan Kumar Bhunia}\textsuperscript{1} \\ \href{https://aneeshan95.github.io/}{Aneeshan Sain}\textsuperscript{1} \hspace{.2cm}  \href{http://www.pinakinathc.me/}{Pinaki Nath Chowdhury}\textsuperscript{1}\hspace{.2cm}
\href{https://www.surrey.ac.uk/people/tao-xiang}{Tao Xiang}\textsuperscript{1,2}\hspace{.2cm} \href{https://personalpages.surrey.ac.uk/y.song/}{Yi-Zhe Song}\textsuperscript{1,2} \\
\textsuperscript{1}SketchX, CVSSP, University of Surrey, United Kingdom.  \\
\textsuperscript{2}iFlyTek-Surrey Joint Research Centre on Artificial Intelligence.\\
{\tt\small \{h.bandyopadhyay, s.koley, a.das, a.bhunia, a.sain, p.chowdhury, t.xiang, y.song\}@surrey.ac.uk\vspace{-1cm}}
}

\newcommand\myfigure{
\centering
\captionsetup{type=figure}
    \includegraphics[width=\textwidth]{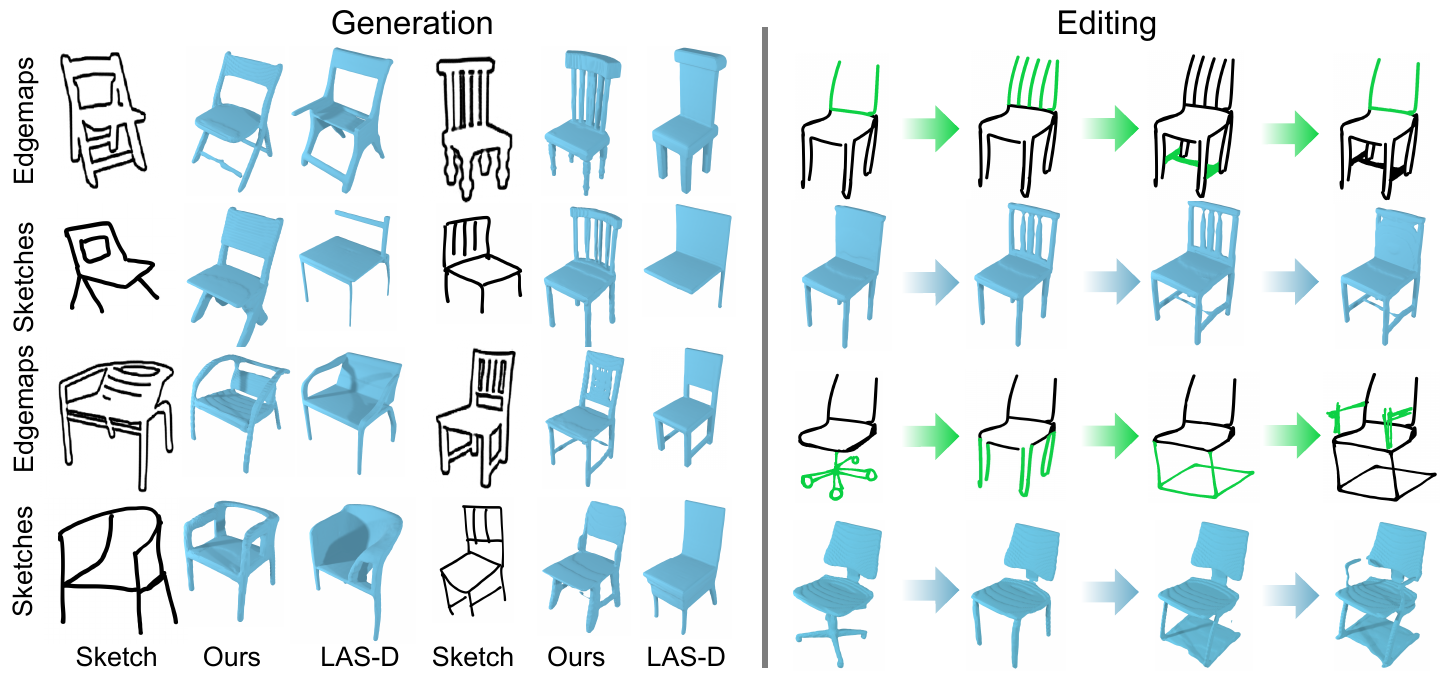}
    \vspace{-0.7cm}
\captionof{figure}{ 
Unlike prior methods (LAS-D \cite{zheng2023lasdiff}), our sketch-based shape generation algorithm generalises to abstract doodles without training on human sketches. Thanks to our part-disentangled sketch and shape representations, we exhibit (a) fine-grained correspondence between sketches and generated shapes, allowing us to (b) perform highly localised shape edits through edits in sketches.}
\label{fig:teaser}
\vspace{+0.2cm}
}

\begin{document}
\maketitle

\begin{abstract}
\vspace{-0.5cm}
In this paper, we democratise 3D content creation, enabling precise generation of 3D shapes from abstract sketches while overcoming limitations tied to drawing skills. We introduce a novel part-level modelling and alignment framework that facilitates abstraction modelling and cross-modal correspondence. Leveraging the same part-level decoder, our approach seamlessly extends to sketch modelling by establishing correspondence between CLIPasso edgemaps and projected 3D part regions, eliminating the need for a dataset pairing human sketches and 3D shapes. Additionally, our method introduces a seamless in-position editing process as a byproduct of cross-modal part-aligned modelling. Operating in a low-dimensional implicit space, our approach significantly reduces computational demands and processing time. 
\end{abstract}

\vspace{-0.4cm}
\section{Introduction}
\label{sec:intro}
\vspace{-0.2cm}
We envisage a world where 3D content creation is democratised, granting everyone the liberty to freely create and modify 3D shapes. This shared vision has spurred collective efforts, initially centred on using text as a condition for 3D shape creation~\cite{liu2023meshdiffusion, li2023diffusion}. However, the pivotal challenge arises when delving into fine-grained \cite{bhunia2020sketch, pang2019generalising, sain2023clip} creation -- text can be ambiguous, lacking the nuanced cues essential for precise idea conveyance. This is where sketches step in as a complementary input modality \cite{chowdhury2023scenetrilogy, sangkloy2022textSketch, song2017textSketch, bandyopadhyay2024xai} , inherently and accurately capturing users' intent -- the promise being ``what you sketch is what you get''.

All that stands in the way of fulfilling that promise is you and your sketch, or the lack of it -- ``I can't sketch'', I hear you say! Prior works \cite{zheng2023lasdiff, guillard2021sketch2mesh,gao2022sketchsampler, binninger2023sens} in this space do generate shapes given rough sketches, but the issue lies in the resulting output -- a poorly drawn sketch yields a deformed 3D shape where your lack of drawing skills is \textit{accurately} reflected -- ``what you sketch is \textit{literally} what you get'' (see \cref{fig:teaser}). In this paper, we aim to democratise sketch-to-3D creation \footnote{\href{https://hmrishavbandy.github.io/doodle23d/}{https://hmrishavbandy.github.io/doodle23d/}} by definitively addressing this very point -- for the first time, enabling your \textit{abstract} (``half-decent'') sketches to generate \textit{precise} 3D shapes, all without any specific paired human sketch and 3D shape data!

Navigating these challenges is no small feat. First, sketches transcend mere edge maps \cite{xu2022deep}; they embody subjective abstract forms that have been proven to be difficult to model \cite{vinker2022clipasso, koley2023picture}. Second, the hurdle lies in injecting a fine-grained understanding to capture the inherent details within sketches. Third, the challenge extends to establishing fine-grained cross-modal correspondences \cite{chowdhury2023democratising} between sketches and 3D shapes. And last but not least, we are keen on avoiding the need to collect a dataset of human sketch and 3D pairs \cite{qi2021toward, zhong2020towards} to achieve all said challenges.

Our solution, we think(!), is quite elegant. It revolves around a well-known hypothesis: part-level modelling aids \cite{corona2021smplicit,wei2022learning} in generalising to complex (fine-grained) shapes. In this paper, we not only extend this hypothesis but also demonstrate its applicability to modelling abstraction. The intuition is straightforward -- parts enable more flexibility (as they can move around) in terms of global construction, thus accommodating deformations \cite{qi2021toward} found in human sketches. It is roughly akin to using a single matchstick where you are limited to just that, but by breaking it into pieces, you can create something more meaningful.

Our part-level modelling initiates from a pre-trained auto-decoder \cite{hertz2022spaghetti}, where we invert the decoder to obtain part-disentangled latent representations. However, this step alone is not sufficient. For cross-modal matching and downstream editing, we also require these parts to be aligned, roughly corresponding to the same regions in an object category. To achieve this, we align these parts, ensuring they share the same indices for similar components in the decomposition. Subsequently, we train a generative model, specifically a diffusion pipeline \cite{ho2020denoising, rombach2022high}, on this aligned latent space. Through sampling from the trained network, we can unconditionally generate aligned and part-disentangled latents. These latents can then be decoded \cite{hertz2022spaghetti} into tangible meshes. Note that in contrast to typical voxel-space diffusion models for 3D shape generation \cite{li2023diffusion,zheng2023lasdiff}, our approach involves performing diffusion in the latent space \cite{rombach2022high}. This approach, operating in a low-dimensional implicit space, enables semantic edits and interpolations with significantly reduced computational demands ($ 0.06 \times$params) and processing time ($3 \times$faster) when compared with prior art \cite{zheng2023lasdiff}.

Curious about the role of our solution in sketch modelling? Here is where the elegance of our approach truly stands out! Surprisingly, the same decoder \cite{hertz2022spaghetti} can be conveniently leveraged to achieve part-level modelling for sketches, all without the need to collect any sketch data. This correspondence is established by transferring the shape decompositions onto their corresponding rendered 2D edgemaps. To enhance abstraction modelling, we took an additional step by passing the aforementioned edgemaps through a CLIPasso \cite{vinker2022clipasso} network to produce of human-like sketches before establishing the said correspondence. Finally, on the editing front, having established a part-level cross-modal correspondence, we can effortlessly determine which parts of the 3D shape the in-position sketch is editing. Subsequently, we can condition the generation of each part with individual part representations. 

In summary, our contributions are: \textit{(i)} Empowering abstract sketches to generate precise 3D shapes and execute in-position editing, surpassing limitations associated with drawing skills prevalent in prior methods. \textit{(ii)} Introducing an intuitive part-level modelling and alignment framework that facilitates abstraction modelling and cross-modal correspondence. \textit{(iii)} Conveniently leveraging the same part-level decoder for sketch modelling, achieved by establishing correspondence between CLIPasso edgemaps and projected 3D part regions. \textit{(iv)} Introducing a seamless in-position editing process as a byproduct of our cross-modal part-aligned modelling.

\vspace{-0.2cm}
\section{Related Works}
\noindent \textbf{3D Representations:} 3D shapes have been popularly represented in their \textit{explicit} geometric form via \textit{(i)} point clouds, \textit{(ii)} voxel grids and \textit{(iii)} polygonal meshes. Generally captured by 3D sensors \cite{huang2022dynamic}, point clouds represent the 3D surface of real-world objects \cite{berger2017survey} using coordinate sets \cite{qi2017pointnet} on a 3D space. While representative in nature \cite{wu2019point}, point clouds are typically sparse and are converted to denser representations \cite{hanocka2020point2mesh, berger2013benchmark} like meshes \cite{hanocka2020point2mesh, zeng2022lion} or used as voxel \cite{zhang2022pvt} grids. Voxelised representations of shapes are dense 3D grids with 
filled/unfilled information per grid-cell, forming the most straightforward extension of a pixel in 3D. Voxels rich in 3D spatial data, are thus commonly processed with 3D CNNs \cite{wu20153d} for both representative \cite{wang2017cnn} and generative \cite{zheng2023lasdiff} tasks. Lighter than voxels, meshes represent approximate 3D \textit{surfaces} with \textit{polygon-faces}, requiring only vertices and connecting faces to depict shapes. This allows easy deformation \cite{litany2018deformable, liu2021deepmetahandles} and manipulation by offsetting vertex positions \cite{liu2023meshdiffusion} for 3D generative tasks \cite{bagautdinov2018modeling, litany2018deformable, liu2023meshdiffusion, hanocka2020point2mesh}. 
Explicit 3D geometrical representations are, however, compute-intensive \cite{tatarchenko2017octree, wang2022dual} and hence limited to low grid-resolutions. 

Alleviating this limitation, continuous \textit{implicit} functions $f_\theta$ \cite{park2019deepsdf,hertz2022spaghetti} represent shapes by mapping coordinates to implicit values. Implicit values like \textit{(i)} occupancy \cite{mescheder2018occupancy} and \textit{(ii)} signed distance \cite{park2019deepsdf} represent the coordinate's \textit{presence} within the shape and distance from the surface respectively. Sampling for these implicit values along a template grid \cite{park2019deepsdf} of mesh vertices or voxel coordinates allows us to build tangible shapes in the form of \textit{meshes} and \textit{voxels}.
Thus, having function parameters $\theta$ enables us to obtain explicit shape forms of arbitrary resolution (0.1 or 0.01 etc.) by sampling any coordinate with the known continuous implicit function.
Beyond implicit and explicit forms, part-level representations decompose 3D shapes into parts for part-specific editing \cite{hertz2022spaghetti, yu2023dualcsg} and generation \cite{nash2017shape, hertz2022spaghetti}. Shapes can be hierarchically decomposed without explicit part-level supervision into hyperplanes \cite{chen2020bsp}, quadratic surfaces \cite{yu2022capri}, superquadrics \cite{paschalidou2019superquadrics, paschalidou2020learning}, and GMM (Gaussian Mixture Model) based \cite{hertz2022spaghetti, hertz2020pointgmm} shape-partitions. 

\noindent \textbf{3D Generation:} 3D shapes can be generated by sampling on a learned latent space of implicit functions like signed distance \cite{park2019deepsdf} and occupancy \cite{mescheder2018occupancy}, or adversarially by converting to explicit voxels \cite{zheng2022sdf, kleineberg2020adversarial}, meshes \cite{luo2021surfgen}, or point clouds \cite{kleineberg2020adversarial}. The recent success of denoising diffusion models \cite{ho2020denoising, song2020denoising, rombach2022high} in generating high fidelity images has led to their adoption for diverse generation of explicit 3D structures \cite{zheng2023lasdiff,liu2023meshdiffusion,Luo_Hu_2021}. Denoising randomly sampled gaussians, however, requires multiple iterations resulting in slow and compute heavy inference \cite{lu2022dpm}, particularly in the context of  explicit generation \cite{zheng2023lasdiff}. As a result, generation of high-resolution shapes becomes \textit{impractical}. While generation of implicit functions \cite{nam20223d, chou2022diffusionsdf} or function parameters \cite{erkocc2023hyperdiffusion} with latent diffusion \cite{rombach2022high} is scalable, it limits generation diversity \cite{hui2022neural} and editability. In this work, we leverage the efficiency of implicit representations with part-level decomposition of GMM based models \cite{hertz2022spaghetti} to learn a latent diffusion model in a part-aware implicit space for high resolution shape generation, guiding and editing.

\noindent \textbf{Sketches for 3D:} Shape reconstruction from single-view \cite{choy20163d, fan2017point, xu2019disn} and multi-view \cite{wang2021multi, choy20163d} images has been extensively studied with deep encoder-decoder architectures \cite{choy20163d, fan2017point} as well as through generative modelling with adversarial \cite{dundar2023progressive, zhang2020image} and diffusion pipelines \cite{gu2023learning, liu2023meshdiffusion}. While sketches are sparser \cite{guillard2021sketch2mesh} representations of shapes compared to images, their expressive nature makes them an ideal input modality for interactive tasks like retrieval \cite{qi2021toward, xu2022domain}, modelling \cite{zheng2023lasdiff, binninger2023sens, li2020sketch2cad, li2022free2cad, wang2018learning}, and editing \cite{zheng2023lasdiff, binninger2023sens} of 3D shapes. However, hand drawn sketches are inaccurate in representing objects from pre-defined viewpoints \cite{gryaditskaya2019opensketch} or with consistent style. This makes the task of sketch-to-3D more of a generative \cite{zheng2023lasdiff} one than reconstructive \cite{binninger2023sens}, as one sketch can correspond to multiple plausible shapes. 
Nevertheless, shapes can be constructed both from reconstruction and generation perspectives with deterministic encoder-decoders \cite{binninger2023sens,li2018robust} and denoising diffusion models \cite{zheng2023lasdiff} respectively. In this work, we encode sketches in a fine-grained part-aware representation for conditional shape generation through latent diffusion \cite{rombach2022high}. We explicitly imbibe part-specific knowledge in sketch representations by aligning part-disentangled sketch encodings with underlying shape parts, allowing us to capture fine-grained shape information from the sketch. 

\vspace{-0.2cm}
\section{Proposed Methodology}
\noindent \textbf{Overview:}
The expressive \cite{hertzmann2020line, bandyopadhyay2024INR} nature of sketch as an input modality makes it an excellent choice for spatial control over 3D generative tasks \cite{zheng2023lasdiff}. We aim to perform sketch-conditioned generation of 3D shapes via \textit{(i)} diffusion-based generative modelling on implicit representations of a pre-trained neural implicit decoder \cite{hertz2022spaghetti} and \textit{(ii)} mapping sketches to this implicit space for fine-grained control. Specifically, we represent part decompositions of 3D shapes in a part-disentangled latent space of a pre-trained decoder \cite{hertz2022spaghetti}, and map sketches to this space via similar part-disentangled sketch representations. Evidently, this mapping allows us to \textit{(i)} condition shapes based on \textit{highly abstract} doodles \cite{ha2017neural} that are \textit{unseen} during training, \textit{(ii)} perform fine-grained and localised shape edits by matching edited regions in sketches, and \textit{(iii)} generate a morph \cite{avrahami2023break} of multiple shapes by naive interpolation of sketches.

\vspace{-0.1cm}
\subsection{Baseline Sketch-to-3D Generation:} 

Implicit Neural functions $f:\mathbb{R}^3 \rightarrow \mathbb{R} $ efficiently characterised 3D shapes with the relationship between 3D coordinates $(x,y,z)$ and their implicit values $O = f(x,y,z)$ as occupancy \cite{mescheder2018occupancy} or signed distance \cite{park2019deepsdf}. Neural networks, as universal function approximators \cite{mildenhall2020nerf}, learn these functions for various shapes as a unified function $O_I = f_\theta(I,x,y,z)$, where implicit code $I \in \mathbb{R}^d$ describes a shape uniquely \cite{park2019deepsdf}. A naive way to model an explicit shape on an input sketch $\mathcal{S}$ would be to learn a mapping from sketch to implicit $I$ using a visual encoder $E$ (like ResNet-18 \cite{he2016deep}) and use a pre-trained function $f_\theta$ to sample implicit values as:

\vspace{-0.2cm}
\begin{equation}
    O_I   = f_\theta(E(\mathcal{S}),X)
    \label{eq: eq-base}
    \vspace{-0.05cm}
\end{equation}
where, $X=(x,y,z)$ refers to sampling coordinates in 3D. The explicit shape can then be reconstructed \cite{park2019deepsdf, hertz2022spaghetti} from a uniform grid of coordinates and their implicit values. While such a formulation explores 3D reconstructions based on efficient neural implicit modelling \cite{park2019deepsdf}, it has a few inherent limitations: \textit{(i)} generated explicit shapes lack editability as \textit{discovering} local edits in implicit codes that translate to explicit edits is non-trivial \cite{hertz2022spaghetti}. This is primarily due to a lack of feature disentanglement in the naive implicit codes $I \in \mathbb{R}^{d}$ that learn a global immutable feature per instance. Secondly, mapping of \textit{sketch} $\rightarrow$ \textit{implicits} is a one-to-many task \cite{zheng2023lasdiff} as 2D line drawings lack the \textit{granularity} to perfectly depict a single shape, corresponding to multiple plausible shapes at once. Modelling such a problem as a deterministic one-to-one mapping \textit{i.e.} $ I = E(\mathcal{S})$ is thus ill-posed. Finally, a mapper trained on synthetic edgemap-like sketches $\mathcal{S}$ (from scarcity of line drawings) does not generalise \cite{zheng2023lasdiff} to human-drawn doodles that suffer from geometric and perspective inaccuracies. 

To address these challenges, we \textit{(i)} perform \textit{part-aware} neural implicit modelling (\cref{sec:part_modelling}) by decomposing explicit shapes into $m$ parts as $Z \in \mathbb{R}^{m\times d}$,
and mapping them to global implicit codes $I$. Part-level disentanglement in the latent space allows for part-specific editing by mixing or swapping parts from multiple latents. \textit{(ii)} We learn a diffusion \cite{ho2020denoising} pipeline as a generative model \cite{rombach2022high} on this part-aware latent space to stochastically generate editable shape implicit codes, where we exert \textit{fine-grained} control over shape generation with sketches. \textit{(iii)} To perform local shape edits by editing on sketches (\cref{fig:teaser}), we learn the same disentanglement of parts -- this time to represent sketches, by matching sketch regions with pre-decomposed parts in $Z$. We find, this helps in generalising to freehand doodles, despite training on synthetic sketches.

\vspace{-0.1cm}
\subsection{Part-aware Neural Implicit Shape modelling} \label{sec:part_modelling}

A pre-trained decoder $\mathcal{D}$ \cite{hertz2022spaghetti} decomposes an explicit shape (mesh) $M$  as $Z = \{\omega_i\}_{i=1}^m$ where $\omega_i \in \mathbb{R}^d$, and maps it to implicit codes $I$, to get output shapes.
These part-latents ($Z$) are decoded to form \textit{(i)} structural representations $Z^{p} =\mathcal{D}_p(Z) \in \mathbb{R}^{m \times d}$ and \textit{(ii)} volumetric descriptors $Z^{g} = \mathcal{D}_g(Z) \in \mathbb{R}^{m \times 16}$ where  $\mathcal{D}_p,\mathcal{D}_g$ are fully connected layers in $\mathcal{D}$. While part-structures ($Z^{p}$) are comprised by $d$-dimensional latents, part-volumes ($Z^{g}$) are represented as parametric 3D gaussians, $\mathcal{N}(\mu_i,\Sigma_i)$
(16-parameters \eg $\mu \in \mathbb{R}^3$, $\Sigma \in \mathbb{R}^{3 \times 3},$ etc. \cite{hertz2022spaghetti, hertz2020pointgmm} ) under a global Gaussian mixture model (GMM) with mixing weights $\pi_i$ as $p(X) = \sum_{i=1}^{m} \pi_i \;\mathcal{N}(\mathbf{\mu_i}, \mathbf{\Sigma_i})$. A Gaussian representing a part in $Z$, captures the probability of a sampled point ($X$ in \cref{eq: eq-base}) to belong to that part.
In unsupervised implicit decomposition, part-specific 3D Gaussians represent part-orientation and positions in the 3D space. They \textit{(i)} enable uniform part decomposition and \textit{(ii)} allow explicit part-disentanglement of overlapping and closely-placed parts. Finally, part-structures and Gaussian parameters are concatenated followed by self-attention \cite{vaswani2017attention} to form implicit codes $I$, thus extending \cref{eq: eq-base} as: 

\vspace{-0.2cm}
\begin{equation}
     O_I   = f_\theta(\mathcal{D}(Z),X)
     \vspace{-0.05cm}
    \label{eq: eq-spaghetti}
\end{equation}
Accordingly, the mesh $M$ can be constructed with Marching Cubes \cite{lorensen1998marching} over a grid of implicit values $O_I$.

\vspace{-0.1cm}
\subsection{Part-level Alignment in the INR Latent Space \label{sec:lat_align}}
\begin{figure}
    \centering
    \includegraphics[width=\linewidth, height=3.5cm]{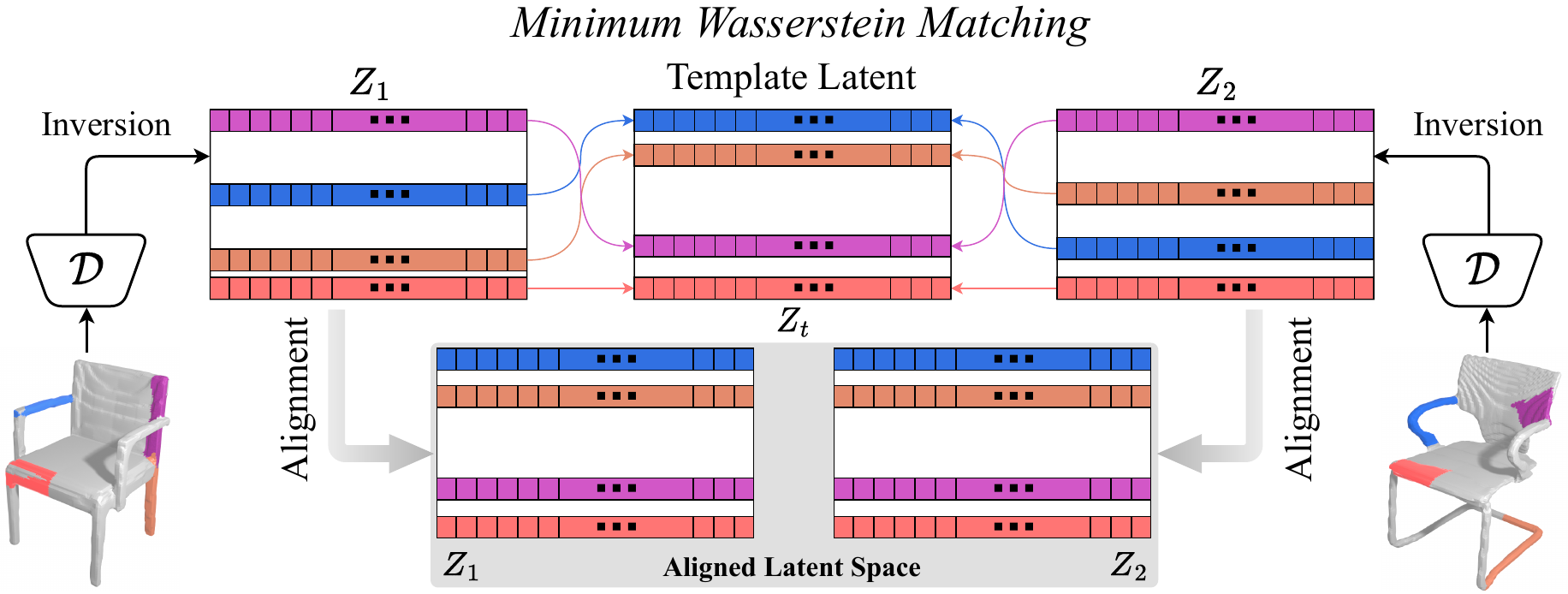}
    \vspace{-0.7cm}
    \caption{Decomposing shapes into latents, we shuffle $m$ part indices of each latent $Z \in \mathbb{R}^{m \times d}$ for minimal Wasserstein distance \cite{kantorovich1960mathematical} with corresponding parts in template latent $Z_t \in \mathbb{R}^{m \times d}$. \\[-0.8cm]}
    \label{fig:alignment}
\end{figure}

Constructing a part-aligned latent representation allows better control for conditional generation and editing \cite{lhhuang2023composer}. Although $Z$ is decomposed into $m$ parts, it lacks part-level alignment across different shapes since the pre-trained decoder $\mathcal{D}$ does not enforce 
positional encoding \cite{hertz2022spaghetti} while using self-attention blocks.
Hence, we align $Z$ such that its part-indices $i \in [1,m]$ correspond (\cref{fig:alignment}) to similar parts (\eg chair's leg) across all shapes. For this, we first pre-compute $Z$ for all shapes by \textit{(i)} using the occupancy values $O_{I}$ and 3D coordinates $X$ to invert our pre-trained decoder $\mathcal{D}$ and implicit function $f_\theta$ as $\mathcal{D}_\phi (.) = \text{Inv}(f_\theta(\mathcal{D}(.),X))$, \textit{(ii)} optimising for $Z = \mathcal{D}_\phi(O_I)$ to match its corresponding implicit code $I = \mathcal{D}(Z)$. Next, we randomly select the part-latents ($Z$) from one shape as \textit{``template part-latents''} $Z_{t} \in \mathbb{R}^{m \times d}$ and align all other $Z$ by minimising the distance between their part-volumes $Z^{g} = \mathcal{D}_g(Z)$ \cite{liu2023meshdiffusion}. Specifically, given the parameters $\{\mu_i,\Sigma_i\}_{i=1}^{m} \in Z^{g}$, we compute the Wasserstein distance ($W_{ij}$) \cite{kantorovich1960mathematical} between two 3D Gaussians -- $\mathcal{N}(\mathbf{\mu_i},\mathbf{\Sigma_i})$ for $i^\text{th}$ part of $Z$ and $\mathcal{N}(\mathbf{\mu_j},\mathbf{\Sigma_j})$ for $j^\text{th}$ part of $Z_{t}$ as:
\vspace{-0.2cm}
\begin{equation}
    W_{i,j}^2 = ||\mu_i -\mu_j||_{2}^{2} + \text{Tr}(\Sigma_i + \Sigma_j -2(\Sigma_i^{\frac{1}{2}}\Sigma_j\Sigma_i^{\frac{1}{2}})^{\frac{1}{2}})
    \vspace{-0.1cm}
\end{equation}
For alignment, we replace the $n^\text{th}$ part of $Z$ with its $\hat{n}^\text{th}$ part that has the minimum Wasserstein distance from $n^\text{th}$ part (same part-indices) of template part-latent $Z_{t}$, as:
\vspace{-0.05cm}
\begin{equation}
    \hat{n} = \argmin_{1 \leq i \leq m} W_{i,n}
    \vspace{-0.2cm}
\end{equation}

\subsection{Unsupervised Part Discovery for Sketches}
\begin{figure}
    \centering
    \includegraphics[width = \linewidth]{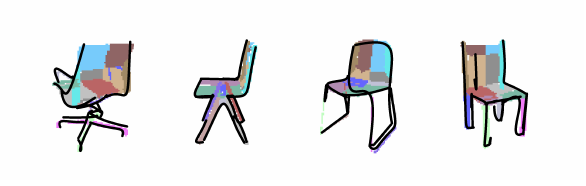}
    \caption{Part-level segmentation maps are created by segmenting 3D shapes into parts with part Gaussians and rendering individual 3D shape parts on synthetic sketches. This segments sketch regions based on shape parts of their corresponding 3D shapes.\\[-0.9cm]}  
    \label{fig:segment}
    \vspace{-0.4cm}
\end{figure}

To enable fine-grained control of shape generation using sketches $\mathcal{S} \in \mathbb{R}^{H \times W}$, we aim to learn an $m$-level part-aligned sketch representation $f_{s}(\mathcal{S}) \in \mathbb{R}^{m \times d_{s}}$. First, we handle the scarcity of paired sketch and 3D shapes \cite{qi2021toward, zhong2020towards}, by rendering 2D projections $\mathcal{I} = \mathcal{R}_{\nu}(M)$ of shape ($M$) \cite{wang2018pixel2mesh} from multiple views $\nu \in \mathcal{V}$ with rendering function $\mathcal{R}_\nu (\cdot)$.
The 2D projections (i.e., images) $\mathcal{I}$ are used to generate synthetic line-drawings $\mathcal{S} = \texttt{sketch}(\mathcal{I})$ as augmented edgemaps \cite{chan2022learning} following \cite{vinker2022clipasso}. 
Next, we learn the $m$-level part-aligned sketch representation $f_{s}(\mathcal{S})$ by discovering parts corresponding to part-latents $Z \in \mathbb{R}^{m \times d}$. Specifically, we use the part-volume parameters $\{ \mu_{i}, \Sigma_{i}, \pi_{i} \} \in Z^{g}$ to represent the probability of a 3D coordinate $X$ belonging to the $i^\text{th}$ part using a GMM with 3D Gaussian $\mathcal{N}(\mathbf{\mu_i} , \mathbf{\Sigma_i})$ and mixing coefficient $\pi_{i}$ as:
\vspace{-0.1cm}
\begin{equation}
    p_i(X) = \pi_i \cdot \mathcal{N}(X|\mathbf{\mu_i}, \mathbf{\Sigma_i})
    \vspace{-0.05cm}
    \label{eq:vol}
\end{equation}
We use $p_i(X)$ to identify parts from shape coordinates $X$ and construct corresponding $m$ 3D segment maps $\{\mathcal{M}_i^\text{3D}\}_{i=1}^m$. Next, we overlay these segmentation maps over synthetic sketches (\cref{fig:segment}) by rendering them from same viewpoint (as sketches) $\nu$ as $\mathcal{M}_{1:m}^{2D} = \{\mathcal{R}_\nu(\mathcal{M}^\text{3D}_i)\}_{i=1}^m$, to construct segmentation maps for sketch regions that correspond to $m$ shape parts. The annotations  $\mathcal{M}_{1:m}^{2D}$ allow us to obtain part-disentangled sketch representations $f_{s}(\mathcal{S})$, aligned with part-latents $Z \in \mathbb{R}^{m\times d}$ of corresponding shapes to enhance control over generation and editing. 

\subsection{Latent Diffusion for 3D Generation \label{sec:latent_diff}}
We use a pre-trained decoder $\mathcal{D}$ to represent shapes in $m$ parts as part-latents $Z \in \mathbb{R}^{m \times d}$. 
$Z$ is aligned across all shapes in a category, such that similar parts have the same part indices \(i \in [1,m]\).
Rendering the \textit{(a)} shape and \textit{(b)} its individual parts on a 2D plane separately, we \textit{(a)} construct sketches from shape boundaries and \textit{(b)} overlay sketch regions with rendered parts (as part-level annotations). Next, we train a diffusion model on part-latents $Z$ for generative modelling, which we condition with part-disentangled sketch representations from part-level annotations.

The diffusion pipeline consists of \textit{(i)} a predefined forward process $q(z_{0:T})$, where noise is added to $z_0$ progressively from $0\rightarrow T$ till $q(z_T)\sim \mathcal{N}(\textbf{0},\textbf{I})$ and a \textit{(ii)} reverse process $p_\theta(z_{T:0})$ where a network estimates denoising conditionals $p_{\theta}(z_{t-1} | z_t)$ for each step $t (<T)$. Specifically, given a noisy sample $z_t = \sqrt{\alpha_t} z_0 + \sqrt{1- \alpha_t} \epsilon$ at time step $t$ with constants $\{\alpha_t\}_{t=0}^{1}$ and $\epsilon \sim \mathcal{N}(\mathbf{0}, \mathbf{I})$, the network estimates the noise $\epsilon$ as $\epsilon_\theta(z_t,t)$. The \textit{simplified} loss is:

\vspace{-0.5cm}
\begin{equation}
\mathcal{L_{\text{LDM}}} =  \mathbb{E}_{z\sim q(z_0), \epsilon \sim \mathcal{N}(\mathbf{0}, \textbf{I}), t} \big[||\epsilon - \epsilon_{\theta}(z_t,t)||_{2}^{2}\big]
\vspace{-0.08cm}
\label{eq: ldm}
\end{equation}

\begin{figure}
    \centering
    \begin{subfigure}[b]{\linewidth}
    \centering
    \includegraphics[width=0.95\linewidth, height=2.5cm]{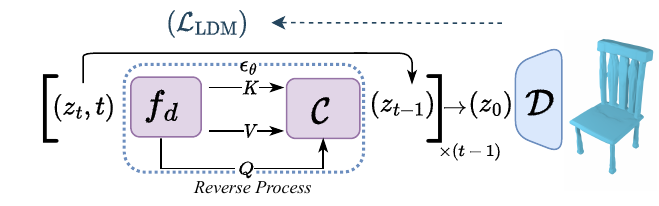}
    \caption{Unconditional Diffusion}
    \end{subfigure}
    \begin{subfigure}[b]{\linewidth}
    \centering
    \includegraphics[width=0.95\linewidth]{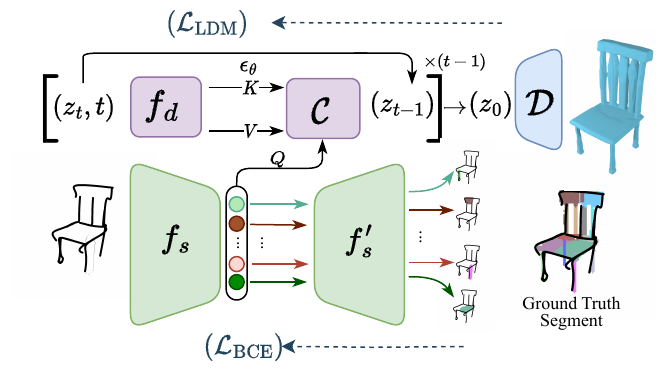}
    \caption{Sketch Conditioned Diffusion}
    \end{subfigure}
    \vspace{-0.7cm}
    \caption{Model overview: (a) The diffusion pipeline denoises latent vector $z_t \in \mathbb{R}^{m \times d}$ to $z_{t-1}$ with fully connected layers $f_d$ and a multi-head attention module $\mathcal{C}$ at time step $t$. After $t=0$, the fully denoised vector $z_0$ corresponds to a generated part-latent $Z$. (b) We encode sketches as part-disentangled representations with encoder $f_s$ by segmenting them into segment maps of individual parts with shared decoder $f_s'$. These sketch representations are fed to the attention module $\mathcal{C}$ as a Query with intermediate diffusion outputs (from $f_d$) as Key-Value pairs.\\[-1cm]}
    \label{fig:network}
\end{figure}

\noindent During inference, $z_T$ is sampled from $\mathcal{N}(\textbf{0},\textbf{I})$ and is iteratively denoised to $z_0$. We learn the underlying distribution of part-latents $Z \in \mathbb{R}^{m \times d}$ as $z_0$ with latent diffusion, where our denoiser $\epsilon_\theta$ consists of fully-connected layers $f_d$ and multi-head attention module $\mathcal{\mathcal{C}}$ (\cref{fig:network}(a)). 

\noindent \textbf{Part-level Sketch Representations: }
For fine-grained control over generated shapes, we use part-disentangled representations $f_s(\mathcal{S}) \in \mathbb{R}^{m \times d_s}$ from sketches $\mathcal{S}$ matching with similar disentanglement in part-latents $Z \in \mathbb{R}^{m\times d}$. For this, we utilise part annotations of sketch regions corresponding to $m$ shape parts and train for segmentation to predict parts as sketch segment-maps. This segmentation is facilitated by encoding sketches as $\eta = f_s(\mathcal{S}) \in \mathbb{R}^{m \times d_s}$ with a ResNet-18 \cite{he2016deep} encoder $f_s$, and decoding to segment-map predictions of individual parts with auxiliary decoder $f_s'$ consisting of upsampling blocks (\cref{fig:network}(b)). To optimize for disentanglement of part features in sketch representations $f_s(\mathcal{S})$, representation of each part of size $\in \mathbb{R}^{d_s}$ is decoded individually with common decoder $f_s'$, making disentanglement indispensable for accurate part-segmentation. 

Importantly through part-aware representations, we align encoding $f_s(\mathcal{S})$ to represent the same explicit parts from sketch $\mathcal{S}$ irrespective of its viewpoint $\nu$. As such, we can naively aggregate information on different parts from multi-view $\{\nu_1,...\nu_n\}$ sketches $\{\mathcal{S}_{\nu_j}\}_{j=1}^{n}$, allowing us to reconstruct the shape more accurately (\cref{fig:variance}) with an aggregated representation $f_s(\mathcal{V}) = \frac{1}{n} \cdot \sum_{j=1}^{n} f_s(\mathcal{S}_{\nu_j})$

\noindent \textbf{Sketch Conditioning: }
Sketches are encoded as $\eta=f_s(\mathcal{S}) \in \mathbb{R}^{m \times d}$ where sketch-features are disentangled into $m$ parts-features as $\{\eta_i\}_{i=1}^{m}$. Part-features have one-to-one correspondence with pre-defined explicit shape parts, and hence part-indexing holds positional significance to uphold this correspondence. We utilise this property for sketch conditioned generation by concatenating sinusoidal positional embeddings of part-indices to both the condition vector $f_s(\mathcal{S})$ from sketch $\mathcal{S}$ and the noised input $z_t$. We then use our multi-head attention block $\mathcal{C}$ as a cross-attention module \cite{vaswani2017attention} by using the sketch conditioning $f_s(\mathcal{S})$ and intermediate diffusion outputs $f_s(\mathcal{S})$ as Query and Key-Value pairs respectively to compute attention as:
\vspace{-0.2cm}
\begin{equation}
\text{Attention} (Q,K,V) = \text{softmax}(QK^T/\sqrt{d_k}) \cdot V
\vspace{-0.05cm}
\end{equation}
with $Q\myeq W^Q \cdot f_s(\mathcal{S}), K\myeq W^K \cdot f_d(z_t,t), V \myeq W^V \cdot f_d(z_t,t)$ 
where $W^Q, W^K, W^V$ represent the projection matrices for Queries, Keys, and Values respectively. The output from $\mathcal{C}$ (used for cross-attention) is projected back to the input dimension with fully connected layers and skip connections. The loss function is extended from \cref{eq: ldm} as:
\vspace{-0.2cm}
\begin{equation}
\mathcal{L_{\text{C}}}\myeq \mathbb{E}_{z \sim q(z_0), \epsilon \sim \mathcal{N}(\mathbf{0}, \textbf{I}), t}\big[||\epsilon - \epsilon_{\theta}(z_t,t, f_s(\mathcal{S}))||_{2}^{2}\big]
\end{equation}

\vspace{-0.1cm}

\subsection{Inference Pipeline}
Inference with the trained diffusion pipeline is performed as a reverse process, by \textit{(i)} randomly sampling $z_T \in \mathbb{R}^{m \times d}$ from the normal distribution $\mathcal{N}(\mathbf{0},\mathbf{I})$ and \textit{(ii)} estimating the denoising conditionals at each step $t:T\rightarrow 0$ as $\epsilon_\theta(z_t,t,f_s\big(\mathcal{S})\big)\in \mathbb{R}^{m \times d}$ with conditioning signal $f_s(\mathcal{S}) \in \mathbb{R}^{m \times d_s}$ from sketch $\mathcal{S}$. The intermediate denoised latent $z_t$ is now denoised to $z_{0}$ with estimated noise iteratively as $z_t \rightarrow z_{t-1} \rightarrow z_{t-2}... \rightarrow z_0$ 
giving us part-latents $Z = z_0 \in \mathbb{R}^{m \times d}$ that can be decoded to obtain implicit codes $I = \mathcal{D}(Z)$ with pre-trained $\mathcal{D}$. Implicit values (occupancies) are sampled with a coordinate grid as \cref{eq: eq-spaghetti} to construct output mesh $M$ with marching-cubes \cite{lorensen1998marching}.

\vspace{-0.1cm}
\section{Experiments}

\noindent Our latent diffusion pipeline is trained on edgemaps \cite{chan2022learning} and synthetic sketches \cite{vinker2022clipasso} from 2D projections of 3D shapes. For generalisation to human sketches, we evaluate primarily on hand-drawn doodles for generative control and diversity.

\begin{figure}[!htbp]
    \centering
    \includegraphics[width=\linewidth]{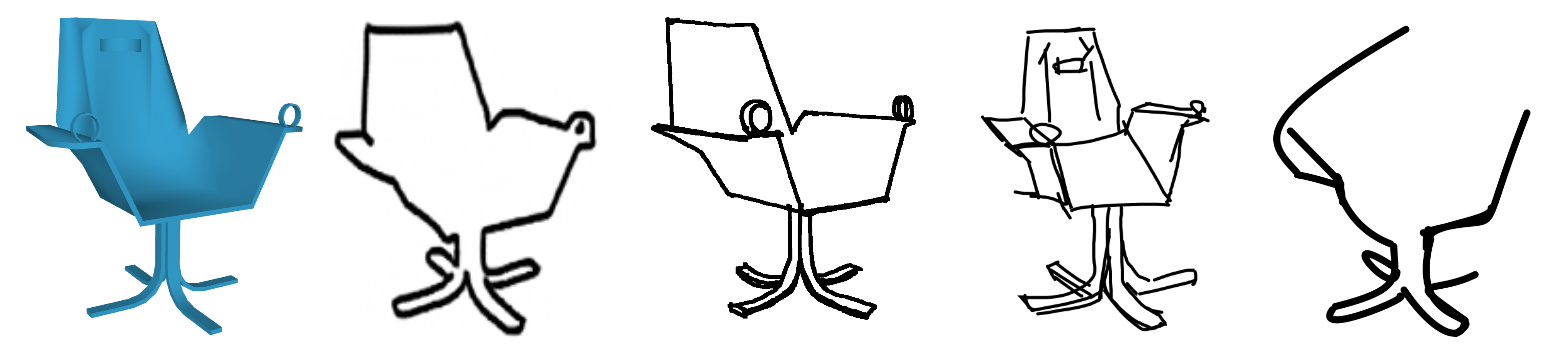}
    \vspace{-0.8cm}
    \caption{From left to right: 3D shape; an edgemap of a 2D render of the shape; corresponding sketches from the ProSketch3D \cite{zhong2020towards} and AmateurSketch3D \cite{qi2021toward} datasets; an abstract CLIPasso \cite{vinker2022clipasso} sketch of the shape.\\[-0.9cm]}
    \label{fig:dataset}
\end{figure}

\noindent \textbf{Datasets}: We use a subset of the ShapeNet \cite{chang2015shapenet} dataset, consisting of $6755$ `chair' shapes \cite{binninger2023sens}.  All shapes are normalised to unit cube and rendered from 6 distinct views  $\nu \in \mathcal{V}$ with azimuthal angles distributed across $[-\pi,\pi)$ and at a constant elevation of $\frac{\pi}{10}$ from a distance of $2.07$ units.  These renders are used to create non photo-realistic renderings \cite{chan2022learning} and abstract sketches using CLIPasso \cite{vinker2022clipasso} similar to \cite{binninger2023sens}  (\cref{fig:dataset}). 
Thus, we obtain a total of $\sim80$K sketch samples synthetically generated from our shape dataset. We invert pre-trained $\mathcal{D}_s$ for these $6755$ shapes, obtaining their latent representations $Z$, which we align and use as ground truth latent codes following the evaluation split in \cite{park2019deepsdf}. Without training on \textit{any} human sketches, we evaluate our model on sketches from the \textit{AmateurSketch-3D} dataset \cite{qi2021toward} consisting of 3000 hand-drawn sketches of $1000$ chairs drawn from viewpoints at angles of $0^{\circ}$, $30^{\circ}$, and $75^{\circ}$. We further evaluate on the \textit{ProSketch} dataset \cite{zhong2020towards} containing $1500$ sketches of $500$ chairs drawn at $0^{\circ}$, $45^{\circ}$, and $90^{\circ}$. Our evaluation set thus consists of \textit{(i)} unseen hand-drawn sketches drawn from \textit{(ii)} unseen azimuthal angles, allowing us to demonstrate the robustness of our algorithm to sketch style and viewpoint perturbation, respectively. Finally, we demonstrate further robustness to abstraction by generating shapes from chair category in the \textit{Quick-Draw!} dataset \cite{ha2017neural} of highly abstract diverse sketches from 15 million people.

\noindent \textbf{Implementation details}: Noisy sample $z_t~\in~\mathbb{R}^{m\times 512}$ is concatenated with sinusoidal positional embeddings ($\in\mathbb{R}^{m \times 224}$) of timestep \(t\), and projected to  $\mathbb{R}^{16\times 128}$. 
Sketch encodings $f_s(\mathcal{S}) \in \mathbb{R}^{16\times64}$ are likewise projected and concatenated with part-index embeddings to $\mathbb{R}^{16\times128}$ forming Query for multi-head attention block $\mathcal{C}$ that generate part-latents $z_0~\in~\mathbb{R}^{m\times 512}$.

\noindent \textbf{Training}: We train the diffusion pipeline with a DDPM solver \cite{song2020denoising} for 1000 timesteps for 10M iterations. We use a learning rate of 1e-4 with the AdamW optimizer and a batch size of 128 on a NVIDIA RTX 3060 Ti. The sketch conditioning module is trained on the same GPU with a learning rate of 2e-4 with an Adam optimizer and a batch size of 64. Training the unconditional diffusion model, the sketch conditioning, and the conditional diffusion pipeline takes 39, 41, and 44 hours on the GPU respectively. 
\begin{figure*}[t]
  \includegraphics[width=\textwidth,trim={1.7cm 0 0 0mm}]{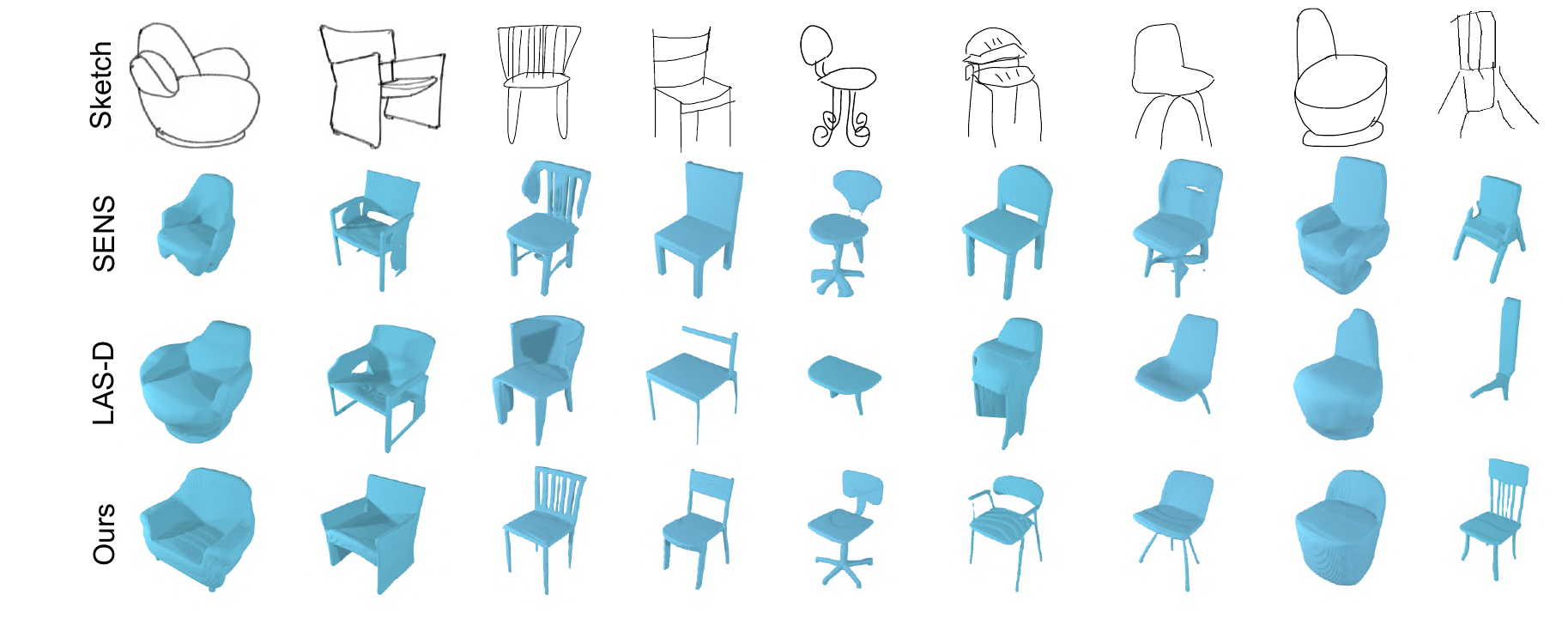}
  \vspace{-1cm}
  \caption{Qualitative comparisons of our method with LAS-D \cite{zheng2023lasdiff} and SENS \cite{binninger2023sens} on sketches of different levels of abstraction from \textit{(i)} highly detailed sketches by artists (first 2 from ProSketch-3D \cite{zhong2020towards}) and \textit{(ii)} sketches by amateurs with perspective distortions (next 6 from AmateurSketch-3D \cite{qi2021toward}) to \textit{(iii)} Highly abstract sketches drawn in $<$20s (last from \textit{Quick-Draw!} \cite{ha2017neural}). While neither LAS-D nor our algorithm has seen hand-drawn doodles during training, SENS\cite{binninger2023sens} was trained on ProSketch-3D sketches.\\[-0.8cm]}
    \label{fig:qual_comp}
\end{figure*}

\noindent \textbf{Competitors}: \textit{(i)} Unconditional shape generation algorithms: \textbf{SDF-StyleGAN} \cite{zheng2022sdf} appropriates StyleGAN2 \cite{karras2020analyzing} from 2D to 3D with 3D convolutions and learns a shape in a voxelised SDF space through locally and globally critical discriminators. 
\textbf{Diffusion-SDF} \cite{chou2022diffusionsdf} learns an MLP-based diffusion model to predict the implicit function for generating a shape. Rather than the `coarse' implicit space, denoising is performed in a `smooth' variational \cite{kingma2013auto} latent space from which an implicit latent code is decoded. \textbf{Mesh-Diffusion} \cite{liu2023meshdiffusion} parameterises meshes with tetrahedral grids, representing them with 3D convolutions and generating shapes from deformations of uniform tetrahedrals. Recent volumetric diffusion models including \textbf{LAS-D} \cite{zheng2023lasdiff}, \textbf{SDFusion} \cite{cheng2023sdfusion} and \textbf{Wavelet-Diffusion} \cite{hui2022neural} perform diffusion in the voxelized occupancy space, SDF space, and SDF-decomposed wavelet space respectively to generate shapes. \textit{(ii)} Conditional generation of shapes has been explored through language prompts \cite{liu2023meshdiffusion}, images \cite{gao2022get3d}, point-clouds \cite{Luo_Hu_2021} and sketches \cite{binninger2023sens, zheng2023lasdiff}. We compare our sketch-based shape generation with \textit{LAS-D} \cite{zheng2023lasdiff} as well as image-based shape generation models like \textit{SDFusion} \cite{cheng2023sdfusion}. We also compare with deterministic sketch to shape reconstruction algorithms. \textbf{Sketch2Model} \cite{zhang2021sketch2model} builds shapes from sketches adversarially with sketch-viewpoint aware generative modelling. \textbf{Sketch2Mesh} \cite{guillard2021sketch2mesh} renders shapes differentially and optimizes the mesh by comparing \textit{(a)} shape silhouette with sketches and \textit{(b)} segmentation masks from shape renders with segmented sketches. \textbf{SENS} \cite{binninger2023sens} learns a Vision-Transformer mapping of sketch patches to latent space of $\mathcal{D}$, forging a \textit{one-to-one} relationship between sketches and corresponding shape implicit codes.
\begin{table}[t!]

\begin{center}
\captionof{table}{\label{tab:cond} Comparison of conditional generation and model efficiency on the AmateurSketch-3D \cite{qi2021toward} and ProSketch-3D \cite{zhong2020towards} datasets (unit for CD here is $10^{-1}$). Performance of SENS \cite{binninger2023sens} not reported for ProSketch-3D as it is in their training set. \\ [-1.7cm] \\ [0.3cm]}
\setlength{\tabcolsep}{3pt}
\scalebox{0.75}{
\begin{tabular}{lcccccc}
\toprule[0.4mm]
& \multicolumn{2}{c}{AmateurSketch-3D} & \multicolumn{2}{c}{ProSketch-3D} & \multicolumn{2}{c}{Inference}\\
\cmidrule(lr){2-3}\cmidrule(lr){4-5}\cmidrule(lr){6-7}
\multirow{-2}{*}{Methods} & CD $\downarrow$ & EMD $\downarrow$& CD $\downarrow$& EMD $\downarrow$& Time $\downarrow$ & Params$\downarrow$\\ \midrule

Sketch2Model \cite{zhang2021sketch2model}  & 0.913 & 0.631 & 1.050 & 0.301 & 1.47s & 85M\\
SketchSampler \cite{gao2022sketchsampler}& 0.615 & 0.537 & 0.582 & 0.240 & 1.64s & 46M \\
Sketch2Mesh \cite{guillard2021sketch2mesh}  & 0.257 & 0.211 & 0.228 & 0.171 & 90s & 9M\\
SENS \cite{binninger2023sens}  & 0.121 & 0.096 & - & - & 3.33s & 177M\\
\midrule
SDFusion \cite{cheng2023sdfusion}  & 0.632 & 0.483 & 0.375 & 0.259 & 25s & 1099M\\
LAS-D \cite{zheng2023lasdiff}  & 0.159 & 0.128 & 0.195 & 0.147& 6s & 767M\\
\midrule
Ours & 0.109 & 0.091 & 0.093 & 0.087 & 2s & 46M\\
Ours-Multi-View & 0.097 & 0.089 & 0.085 & 0.082 & 2.5s & 46M
\\  \bottomrule[0.4mm]
\end{tabular}
}
\end{center}\vspace{-0.5cm}
\end{table}

\noindent \textbf{Comparative Analysis}: We include a qualitative performance analysis of our algorithm with SOTA LAS-D \cite{zheng2023lasdiff} and SENS \cite{binninger2023sens} on different sketch datasets in \cref{fig:qual_comp}. For quantitative evaluation of conditional shape generation, we follow recent works \cite{zheng2023lasdiff} to sample 2048 points on both generated and ground truth meshes and use point-cloud based metrics like \textit{(i)} \textbf{Chamfer Distance} (CD) computed as the squared distances between nearest points in predicted and ground truth point clouds and \textit{(ii)} \textbf{Earth Mover's Distance} (EMD) computed as the average point-to-point distance under a global match of the prediction with the ground truth. From \cref{tab:cond}, our algorithm outperforms previous SOTA LAS-D \cite{zheng2023lasdiff} and SENS \cite{binninger2023sens} in sketch-conditioned generation by $0.0076/0.049$ and $0.0012/0.05$ on average CD/EMD respectively without any training on human sketches. 
For unconditional shape generation (\cref{tab:acc}), we evaluate diversity with shading-image based \textbf{Fr\'echet Inception Distance} (FID) \cite{zheng2022sdf} where images are rendered both from the predicted and ground truth meshes in $20$ uniform views, from which view-specific FIDs are averaged. 
We find that our method outperforms Mesh-Diffusion \cite{liu2023meshdiffusion}, SDF-StyleGAN \cite{zheng2022sdf}, and Wavelet Diffusion \cite{hui2022neural} in FID by $23.57$,$14.16$, and $6.32$ respectively (lower is better) while performing at par with frameworks like LAS-D \cite{zheng2023lasdiff} and Diffusion-SDF \cite{chou2022diffusionsdf}. We also use point-cloud based metrics \cite{achlioptas2018learning, yang2019pointflow} like \textit{(i)} \textbf{Coverage} (Cov.) measuring the fraction of generated shape point clouds that match (based on CD/EMD) with the ground truth shape point clouds, \textit{(ii)} \textbf{Minimum Matching Distance} (MMD) measuring the average minimum matching distances (Chamfer/Earth Mover Distances) between point clouds from generated and ground truth shapes and \textit{(iii)} \textbf{1-Nearest Neighbour Accuracy} (1-NNA) measuring the similarity between generated and ground truth point clouds using a 1-NN classifier. While for Cov. and MMD, a higher and lower value respectively is better, for 1-NNA, the best model should be closest to $50$ \% denoting that the 1-NN classifier is confused whether a sample is real or generated. We evaluate with these metrics on the ShapeNet \cite{chang2015shapenet} dataset where our model performs (\cref{tab:acc}) at par with other diffusion-based 3D generative frameworks.

\begin{table}[t!]
\begin{center}
\caption{\label{tab:acc} Comparison of unconditional shape generation from training on ShapeNet Chairs (unit for CD here is $10^{-1}$).\\[-1cm]}

\setlength{\tabcolsep}{3pt}
\scalebox{0.75}{
\begin{tabular}{lccccccc}
\toprule[0.4mm]
& \multicolumn{2}{c}{COV(\%) $\uparrow$} & \multicolumn{2}{c}{MMD $\downarrow$} & \multicolumn{2}{c}{1-NNA (\%)}\\
\cmidrule(lr){2-3}\cmidrule(lr){4-5}\cmidrule(lr){6-7}

\multirow{-2}{*}{Methods} & CD & EMD & CD & EMD & CD & EMD & \multirow{-2}{*}{FID $\downarrow$} \\ \midrule
SDF-StyleGAN \cite{zheng2022sdf}  & 45.60 & 45.50 & 0.158 & 0.184 & 63.25 & 67.80 & 36.48\\

Diffusion-SDF \cite{chou2022diffusionsdf}  & 65.35 & 59.22 & 0.106 & 0.133 & 51.18 & 54.3 & 21.07\\
Mesh-Diffusion \cite{liu2023meshdiffusion}  & 46.00 & 46.71 & 0.132 & 0.173 & 53.69 & 57.63 & 39.62\\
LAS-D \cite{zheng2023lasdiff}  & 53.76 & 52.43 & 0.138 & 0.175 & 64.53 & 65.15 & 20.45\\
Wavelet-Diffusion \cite{hui2022neural}  & 52.88 & 47.64 & 0.133 & 0.173 & 61.14 & 66.92 & 28.64\\
\midrule
\textbf{Ours} & \textbf{63.39} & \textbf{61.2} & \textbf{0.103} & \textbf{0.149} & \textbf{54.25} & \textbf{57.12} & \textbf{22.32}\\  
\bottomrule[0.4mm]
\end{tabular}
}
\end{center}\vspace{-0.55cm}
\end{table}

\begin{figure}
\vspace{0.5cm}
\RawFloats
\begin{floatrow}
  \begin{minipage}[t]{.57\linewidth}
    \centering
    \includegraphics[width=\linewidth]{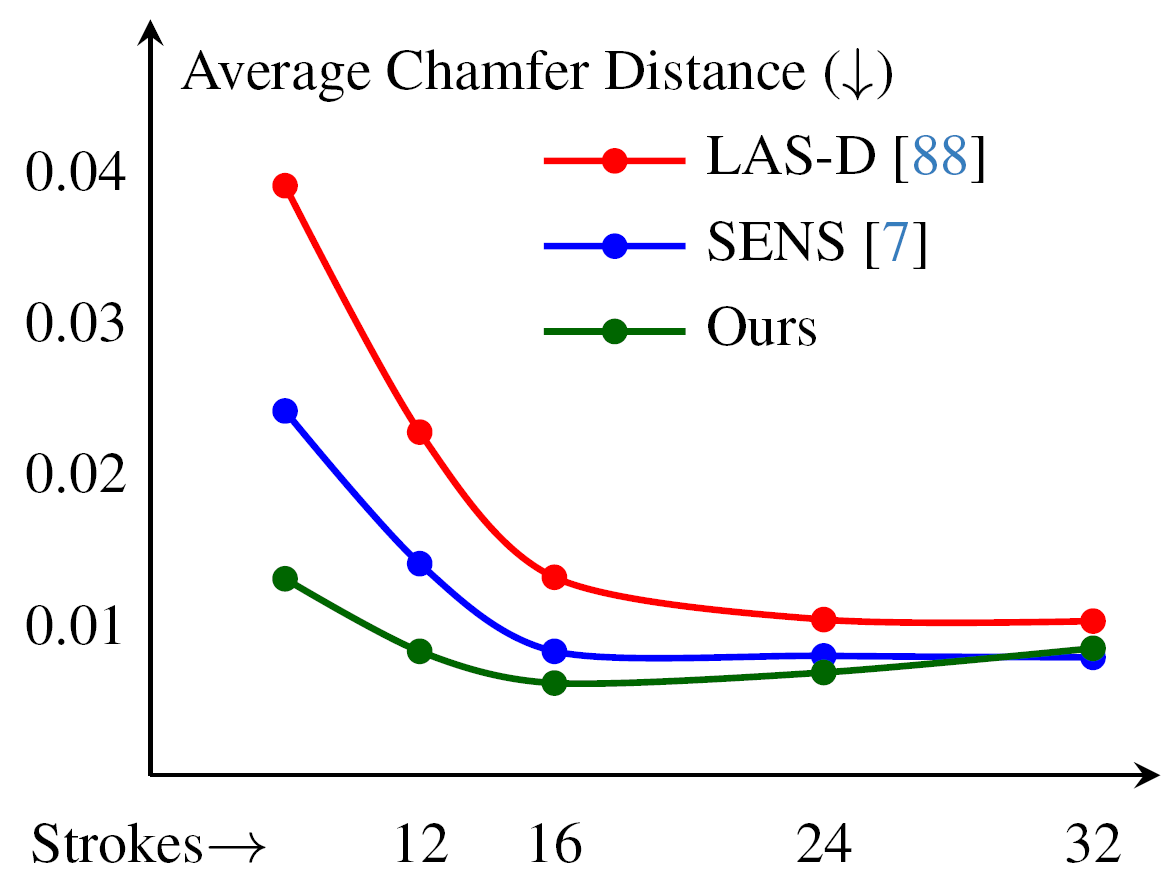}
    \vspace{-0.7cm}
    \caption{Drop in quality of generated shapes in LAS-D \cite{zheng2023lasdiff} and SENS \cite{binninger2023sens} from increasingly abstract sketches generated with CLIPasso \cite{vinker2022clipasso}.}
    \label{fig:enter-label}
  \end{minipage}
  \hspace{3mm}
  
  \begin{minipage}[t]{.39\linewidth}
    \vspace{-3.7cm}
    \scriptsize
    \begin{tabular}{lc}
    \toprule
         Methods &CD $\downarrow$ \\
         \midrule
         \textit{Ours} &0.109\\
         \midrule
         w/o \\
         alignment & 0.171\\
         part-disentangled \\
         sketch encodings & 0.162\\
         \midrule
         with \\
         1 Att$^{\text{n}}$ layer  &0.127\\
         2 Att$^{\text{n}}$ layers &0.114\\
         4 Att$^{\text{n}}$ layers &0.110\\
        \bottomrule
    \end{tabular}
    \vspace{0.15cm}
    \captionof{table}{Ablative studies on the AmateurSketch-3D dataset \cite{qi2021toward} (unit for CD here is $10^{-1}$). \\[-1.2cm]
        \label{tab:ablation}}
  \end{minipage}
  \end{floatrow}
  \vspace{-0.45cm}
\end{figure}

\noindent \textbf{Abstraction robustness}: We use CLIPasso \cite{vinker2022clipasso} to simulate sketches from edgemaps of $100$ random rendered shapes with $8$, $12$, $16$, $24$, and $32$ strokes to represent increasing abstraction levels following their assumption \cite{vinker2022clipasso} of abstraction as a function of number of strokes. Sketches of varying abstraction levels are then used to generate 3D shapes whose Chamfer Distance (CD) from the ground truth shape is plotted in \cref{fig:enter-label}. Despite both our method and SENS \cite{binninger2023sens} being trained on CLIPasso sketches, we find that SENS performs worse at lower abstraction levels than our algorithm, with the average CD increasing from $0.0078$ to $0.0241$ compared to our CD increase ($0.0084$ to $0.0130$). Algorithms like LAS-D \cite{zheng2023lasdiff} train on edgemaps only, resulting in a lower performance (CD increases from $0.0102$ to $0.0390$).

\begin{figure}
\begin{center}
    \centering
    \includegraphics[width=\linewidth]{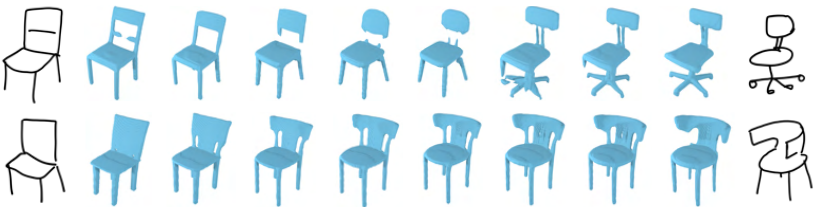}
    \captionof{figure}{Generated shapes can be smoothly morphed into one another by simple interpolation of sketch representations.
    \\[-1.1cm]}
    \label{fig:editing}
    \vspace{-0.6cm}
\end{center}
\end{figure}

\noindent \textbf{Shape editing}:
We explore the editability of generated shapes through \textit{(i)} edits in the input sketch and \textit{(ii)} interpolation between shapes. Representing sketches through local part-aware encodings allows us to perform local edits \cite{hertz2022spaghetti} in the shape without disturbing the global structure, with stroke-level sketch edits. Given a sketch edit $\mathcal{S} \rightarrow \mathcal{S}'$, we identify parts affected in the edit, by the Euclidean distance between part encodings. Next, we generate corresponding part-latents $Z$ and $Z'$, and then replace part information in $Z$ corresponding to edited parts in $S'$ with respective part information from $Z'$. As observed in \cref{fig:teaser} (right), we are able to ensure local edits while maintaining fine-grained control over generated shapes. \cref{fig:editing} demonstrates edits by morphing shapes into one another via interpolation of sketch representations only.

\begin{figure}
\begin{center}
    \centering
    \captionsetup{type=figure}
    \includegraphics[width=\linewidth]{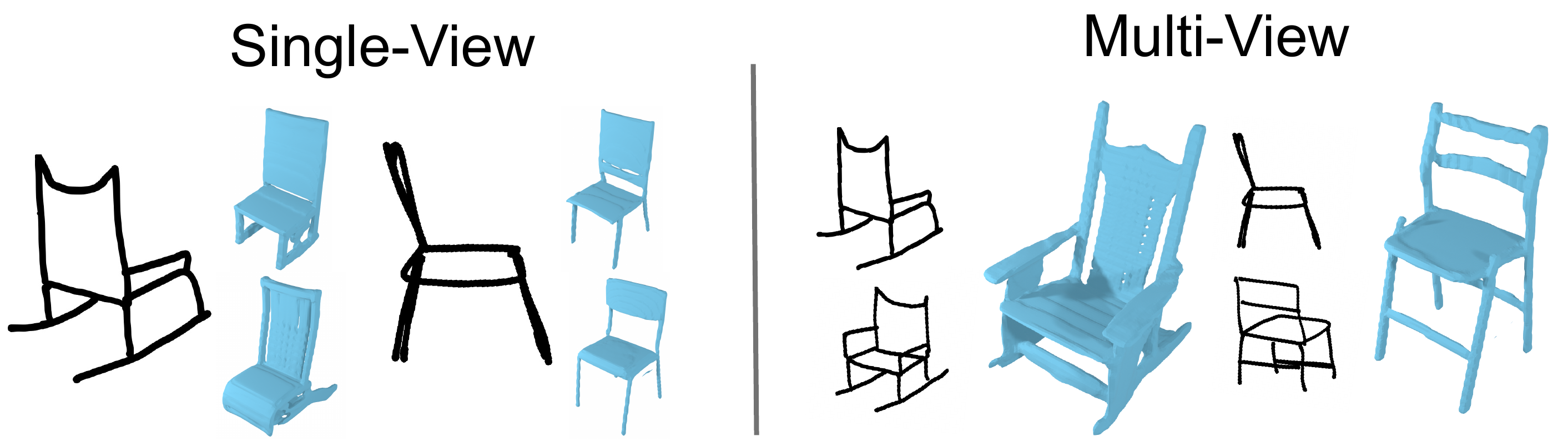}
    \vspace{-1cm}
    \captionof{figure}{Ambiguity in the input sketch, owing to occlusion from specific views or poor quality sketches leads to (slight) variance in generated shapes that can be fixed with multi-view sketches.\\[-1cm]}
    \label{fig:variance}
\end{center}
\end{figure}

\noindent \textbf{Ambiguity in Sketches}: 
Single-view sketches of 3D shapes suffer from ambiguity in representing complex structures due to \textit{(i)} the sparse nature of line drawings and \textit{(ii)} the limited representative power of a single viewpoint . Deterministic reconstruction of shapes from sketches is thus inherently \textit{biased} as they selectively form one-to-one sketch-to-shape correspondences in a one-to-many setting. Recognising this ambiguity, we demonstrate shape variations 
from single-view sketches from AmateurSketch-3D in \cref{fig:variance}. 
To reduce ambiguity in generated shapes, we perform multi-view sketch to shape reconstruction by simply aggregating part representations from multiple views. This increases shape reconstruction accuracy (\cref{fig:variance}) in AmateurSketch-3D and ProSketch-3D sketches, dropping CD/EMD from $0.0109/0.093$ to $0.0097/0.085$ respectively (\cref{tab:cond}).

\noindent \textbf{Ablation Experiments}: We perform ablative studies to analyse the contribution of individual elements in our proposed approach. Using Chamfer Distance as a metric on the AmateurSketch-3D \cite{qi2021toward} dataset, we summarise our results in \cref{tab:ablation}. We find a increase in CD of \textit{(i)} $0.0062$ on removing latent alignment (\cref{sec:lat_align}), \textit{(ii)} of $0.0053$ on replacing part-aware sketch encodings (\cref{sec:latent_diff}) with naive ImageNet features, and \textit{(iii)} by $0.0018/0.0005/0.0001$ with $1/2/4$ multi-head attention layers respectively. Removing part-aware sketch features, in particular (\cref{tab:ablation} - w/o part-disentangled sketch encodings), results in much worse generalisation with $+0.0053$ CD on human-drawn sketches compared to synthetic sketches ($+0.0031$ CD).

\noindent \textbf{Time and memory constraints}: Towards practical generation of 3D  shapes, we evaluate the compute constraints of our algorithm against conditional shape generative networks like LAS-D, SENS, and SDFusion (\cref{tab:cond}). Our model not only outperforms conditional networks in parameter count, but is also lighter than unconditional generative models (which do not have a conditioning network) like Diffusion-SDF ($123$M), and Wavelet-Diffusion ($99$M). As a lightweight network, our model naturally has minimal inference time in comparison to SOTA generative networks.  

\vspace{-0.25cm}
\section{Conclusion}
\vspace{-0.1cm}

We present a latent diffusion pipeline to generate precise 3D shapes from abstract sketches with part-disentangled sketch representations. We demonstrate fine-grained control over generated shapes with hand-drawn doodles on a variety of abstraction levels -- from highly abstract \textit{Quick-Draw!} sketches to artistic ProSketch-3D diagrams. Our generated shapes can be automatically edited with fine-grained edits on the conditioning sketch and can be further improved with sketches from additional viewpoints by a simple aggregation of sketch features. Finally, we demonstrate our pipeline to be much more efficient than SOTA 3D generative models.

{
    \small
    \bibliographystyle{ieeenat_fullname}
    \bibliography{arxiv_CR}
}

\clearpage

\appendix

\clearpage
\setcounter{page}{1}

\makeatletter
  \@namedef{figure}{\killfloatstyle\def\@captype{figure}\FR@redefs
    \flrow@setlist{{figure}}%
    \columnwidth\columnwidth\edef\FBB@wd{\the\columnwidth}%
    \FRifFBOX\@@setframe\relax\@@FStrue\@float{figure}}%
\makeatother

\maketitlesupplementary

\section{Segmenting Hand-Drawn Sketches} 
\begin{figure}[!htbp]
    \centering
    \includegraphics[width=0.9\linewidth]{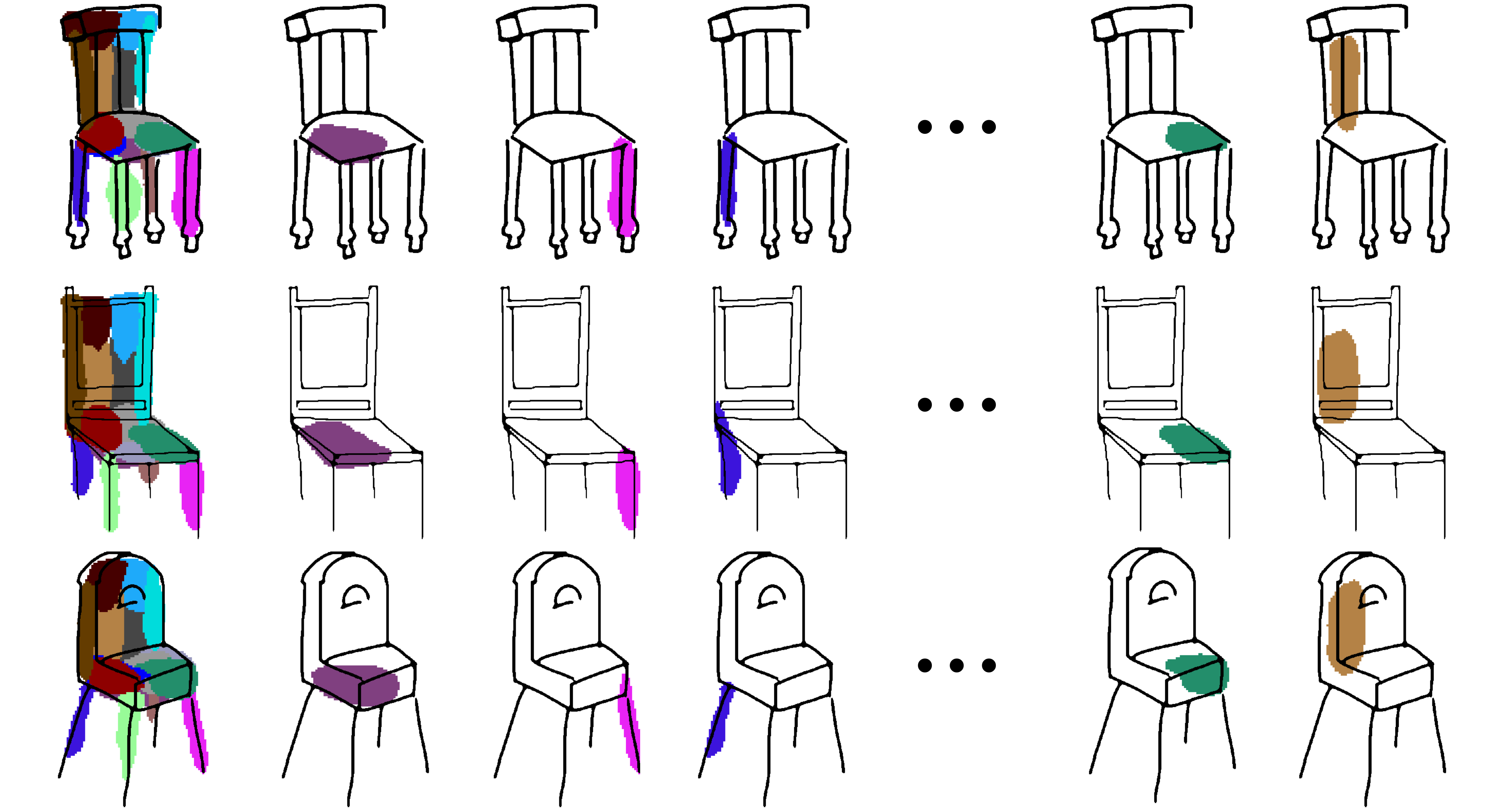}
    \caption{Segmentation results (right) with predicted semantic segmentation map for all $16$ parts (left).}
    \label{fig:seg_extra}
\end{figure}

We demonstrate the generalisation of sketch-based shape generation to hand-drawn sketches after being trained on synthetic sketches only (\cref{fig:qual_comp}). To further explore this generalisation, we include qualitative results from our auxiliary segmentation task on hand-drawn sketches from the AmateurSketch-3D dataset \cite{qi2021toward} in \cref{fig:seg_extra}. We note that despite being trained on synthetic sketches, we can generalise and segment hand-drawn sketches fairly accurately.

\section{Further Details on Shape Decomposition}
Here, we clarify Sec.~\textcolor{red}{3.2}, elaborating on the decomposition of shapes (as meshes) into part-latents $Z$. To represent a ground truth shape $M$ with decoder $\mathcal{D}$, \textit{(i)} the ground-truth occupancy values for coordinates $X = (x,y,z) $ in and around the shape are recorded along with the coordinates themselves. \textit{(ii)} The decoder $\mathcal{D}$ is trained to decode a randomly initialised part-latent $Z$ to an implicit code as \(I=\mathcal{D}(Z)\). \textit{(iii)} Finally $I$ is used in implicit function $f_\theta$ to \textit{predict} occupancy values for known coordinates $X$, as $O_I = f_\theta(I,X)$.

During pre-training, $\mathcal{D}, Z,$ and $f_\theta$ are optimized together with binary cross entropy loss ($\mathcal{L}_\text{BCE}$) against the recorded (ground truth) occupancy values. 

The decomposition of shape $M$ occurs through its representation as part-latent $Z \in \mathbb{R}^{m \times d}$ using the decoder $\mathcal{D}$. Ideally, after disentanglement each latent code $\omega_i$ in part-latent $Z = \{\omega_i\}_{i=1}^m$, sufficiently and independently represents individual components of shape $M$, thus successfully breaking down $M$ into $m$ parts represented as $\{\omega_i\}_{i=1}^m$. 
This disentanglement of part-latents is necessary for independent representation of shape parts. However, $\mathcal{L}_{BCE}$ is not enough for this disentanglement, as it only encourages the final output shape to match shape $M$, thus ignoring \textit{part-level} correspondence. 

To optimize for disentanglement, part-latent $Z$ is projected to part structural representation $Z \rightarrow Z_p$ and part volumetric descriptor $Z \rightarrow Z_g$. Particularly important for this representation, each part's volumetric descriptor is a parametric 3D Gaussian that captures the probability of a 3D coordinate $X$ belonging to that part. This establishes a relationship between coordinates in the 3D space and part-latent $Z$, thereby representing the volume of each part in 3D. For decomposition and disentanglement respectively, this relation of part and 3D coordinates is pivotal for \textit{(i)} dissipating 3D Gaussians $Z_{g}$ over the entire shape volume ($M$) and \textit{(ii)} specifically disentangling overlapped, or closely-placed parts (as information is commonly entangled here \cite{hertz2022spaghetti}), by computing distance between part-Gaussians in 3D.

\section{Shape Interpolation}
\begin{figure}[!htbp]
    \centering
    \includegraphics[width=0.9\linewidth]{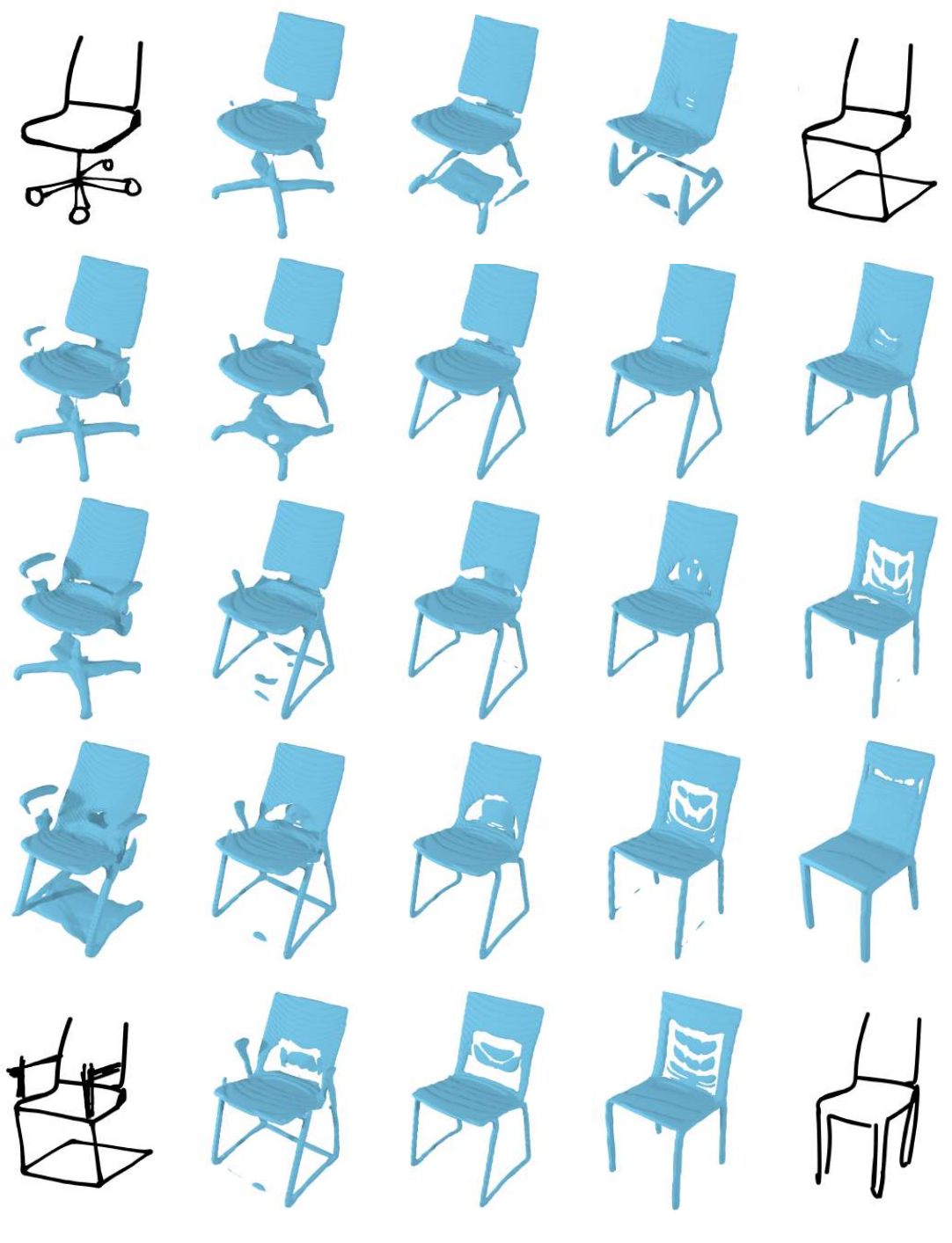}
    \caption{Interpolating shapes among four different sketches (at corners) from left to right and top to bottom.}
    \label{fig:interp_extra}
\end{figure}

\section{Generation of Shapes from Other Categories} In addition to chairs (\cref{fig:qual_comp}), we perform sketch-based 3D generation of \textit{airplanes}, \textit{tables}, \textit{rifles} and \textit{cars}. 

\begin{figure}[!htbp]
    \centering
    \includegraphics[width=0.85\linewidth]{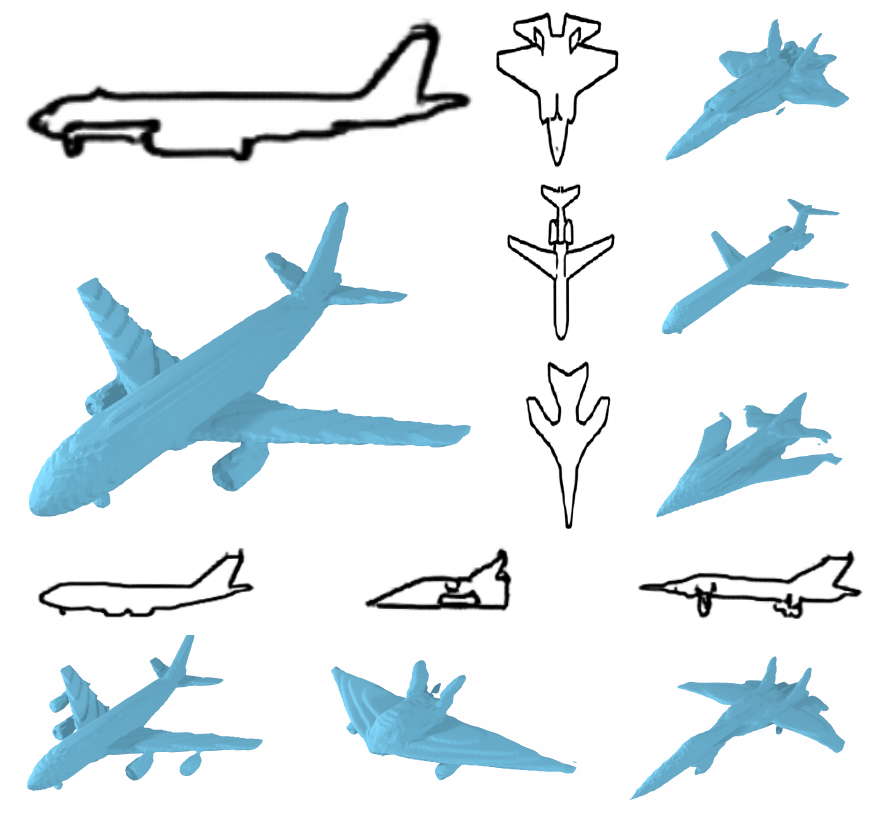}
    
    \includegraphics[width=0.85\linewidth]{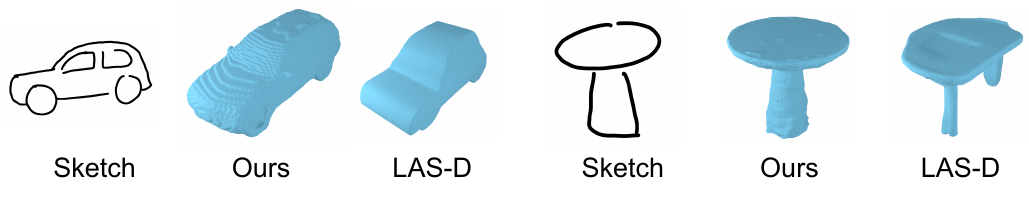}
    \vspace{0.4cm}
    
    \label{fig:seg_add1}
\end{figure}

\section{Additional Qualitative Results (Chairs)}

\begin{figure}[!h]
    \centering
    \vspace{-0.4cm}
    \includegraphics[width=0.9\linewidth]{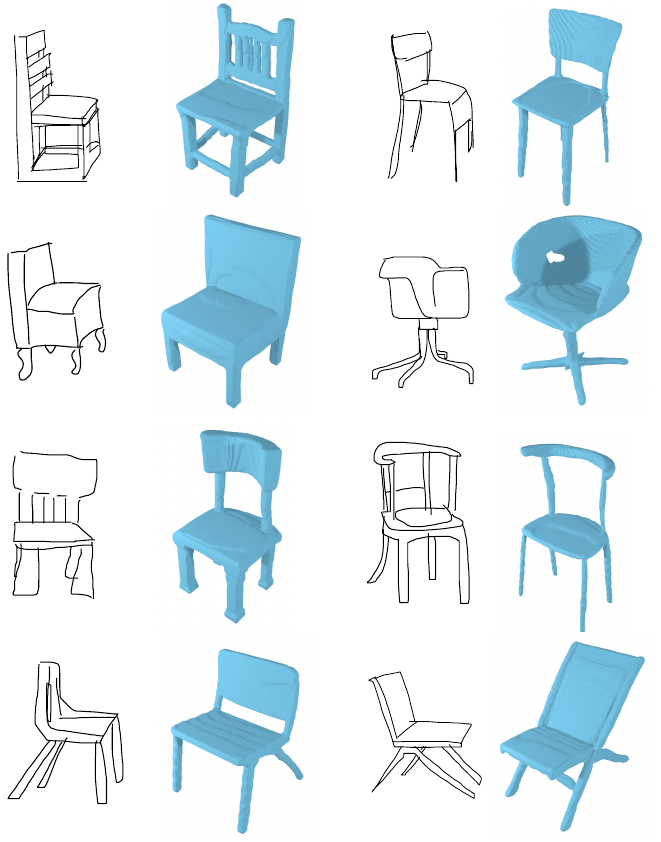}
    \vspace{-0.9cm}
    
    \label{fig:seg_add2}
\end{figure}

\begin{figure}[!h]
    \centering
    \vspace{1.3cm}
    \includegraphics[width=0.95\linewidth]{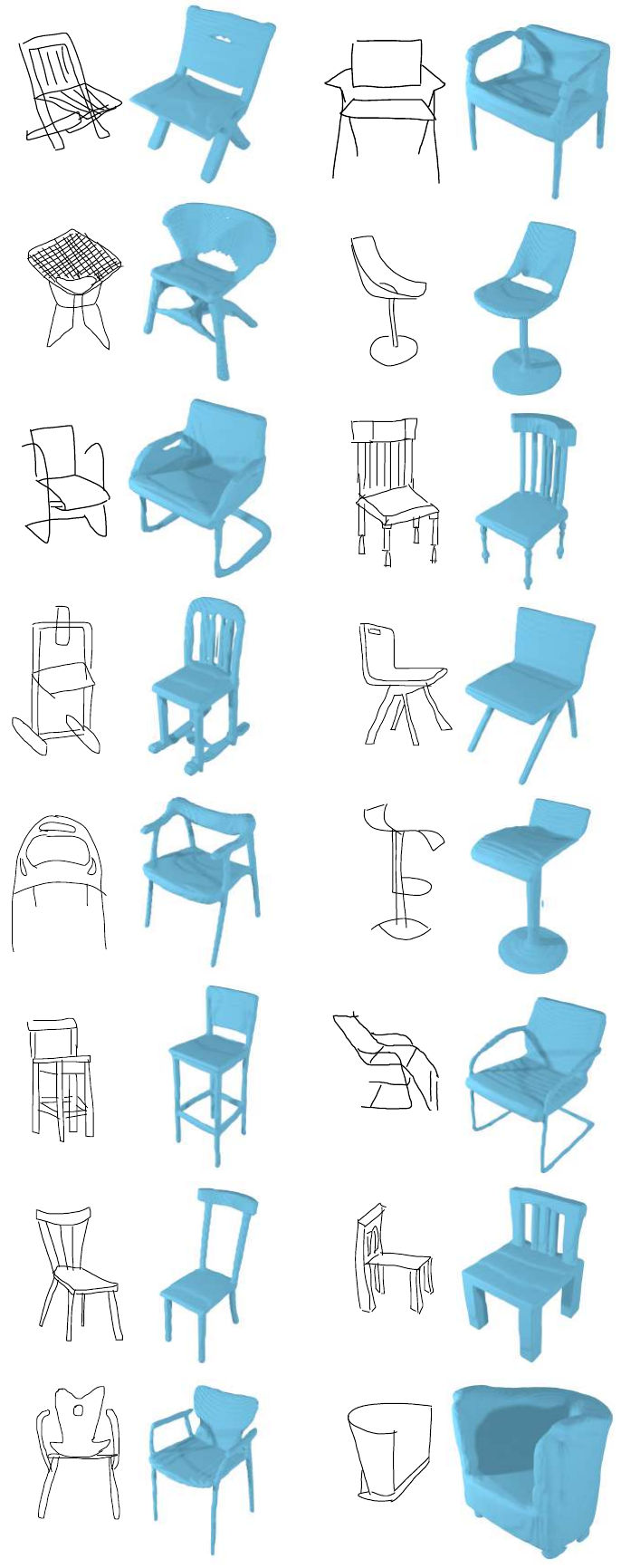}
    \label{fig:seg_add3}
\end{figure}

\clearpage

\section{Model Response to extreme inputs:}

\includegraphics[width=\linewidth]{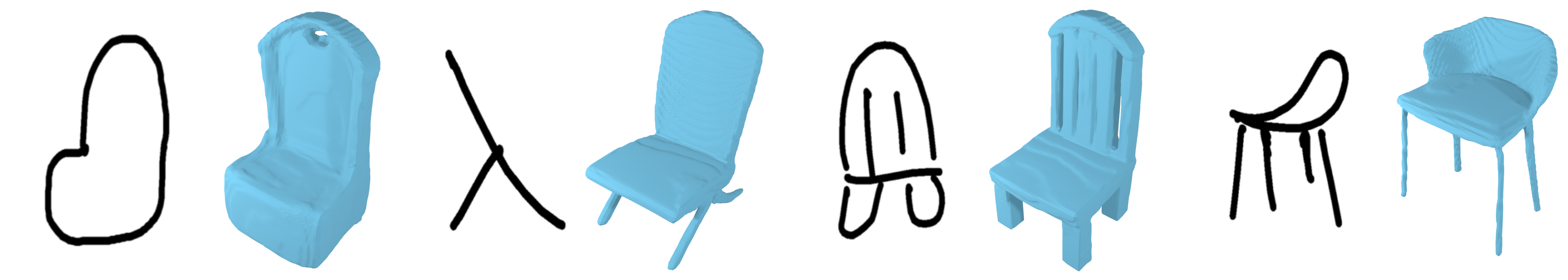}\vspace{-0.2cm}

\hspace{-0.4cm}\includegraphics[width=\linewidth]{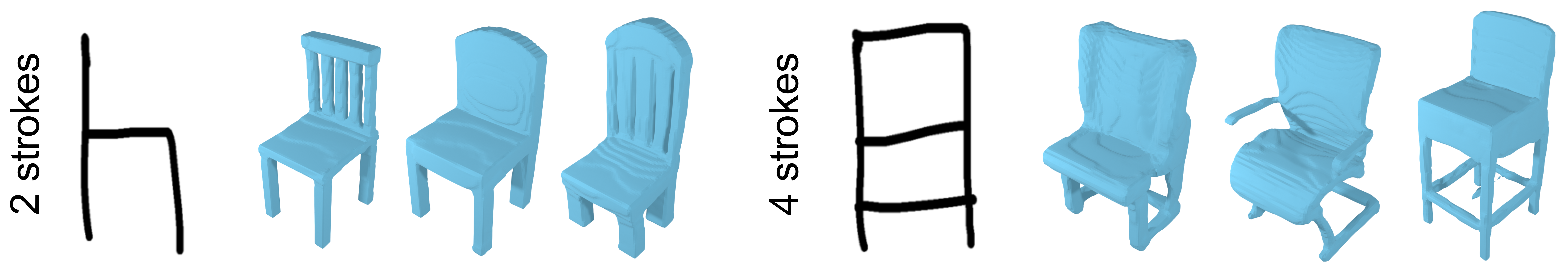}

\section{User feedback for generated shapes:} We build an internal demo using Gradio for shape generation and editing, and ask $30$ users to draw $10$ sketches each on the demo-canvas and rate \textit{(i)} the generated shapes on score from $1$$\to$$5$ (bad $\to$ excellent) based on how they match their expectation. We then ask the same users to edit their sketches and rate the edited shape based on \textit{(ii)} localisation of edits, and \textit{(iii)} quality of details added, using scores ($1\rightarrow5$). Users reported a mean opinion score (MOS) of $4.17/4.00$ for Ours/LAS-D generation quality, $4.91/4.35$ for localisation, and $4.30/3.90$ for quality of edits. We also obtained (iv) a satisfaction score of $4.37/3.55$ for Ours/LAS-D from the same users based on generation speed, shape quality, consistency, and resolution by rating from $1\rightarrow5$. None of them were linked to the project to prevent conflicts.

\end{document}